# STNN-DDI: A Substructure-aware Tensor Neural Network to Predict Drug-Drug Interactions

Hui Yu[1,*], Shi-yu Zhao[1] and Jian-yu Shi[2,*]

[1]School of Computer Science, Northwestern Polytechnical University, Xi'an 710072, China,
[2]School of Life Sciences, Northwestern Polytechnical University, Xi'an 710072, China.
Corresponding authors: Hui Yu, huiyu@nwpu.edu.cn; Jian-yu Shi, jianyushi@nwpu.edu.cn

**Abstract**
Computational prediction of multiple-type drug-drug interaction (DDI) helps reduce unexpected side effects in poly-drug treatments. Although existing computational approaches achieve inspiring results, they ignore that the action of a drug is mainly caused by its chemical substructures. In addition, their interpretability is still weak. In this paper, by supposing that the interactions between two given drugs are caused by their local chemical structures (substructures) and their DDI types are determined by the linkages between different substructure sets, we design a novel Substructure-ware Tensor Neural Network model for DDI prediction (STNN-DDI). The proposed model learns a 3-D tensor of (substructure, interaction type, substructure) triplets, which characterizes a substructure-substructure interaction (SSI) space. According to a list of predefined substructures with specific chemical meanings, the mapping of drugs into this SSI space enables STNN-DDI to perform the multiple-type DDI prediction in both transductive and inductive scenarios in a unified form with an explicable manner. The comparison with deep learning-based state-of-the-art baselines demonstrates the superiority of STNN-DDI with the significant improvement of AUC, AUPR, Accuracy, and Precision. More importantly, case studies illustrate its interpretability by both revealing an important substructure pair across drugs regarding a DDI type of interest and uncovering interaction type-specific substructure pairs in a given DDI. In summary, STNN-DDI provides an effective approach to predicting DDIs as well as explaining the interaction mechanisms among drugs. Source code is freely available at https://github.com/zsy-9/STNN-DDI.

## 1 Introduction

In the treatment of human complex disease, co-prescription of drugs has become a common therapy[1] because individual drugs often fail to meet clinical needs. However, two or more drugs are taken together would induce pharmacokinetic or pharmacodynamic changes[2][3], referred to as drug-drug interactions (DDIs), which may trigger unexpected adverse drug events[4]. DDI-triggered adverse effects are harmful even dangerous to patients[5]. Therefore, the detection of potential DDIs before making co-prescriptions has been gaining attention from biologists, pharmacologists, and clinicians. Due to the enormous combinational number of drugs, traditional methods (e.g., in-vitro approaches and clinical trials) are time-consuming, low-efficiency, and costly[6]. In recent years, computational methods have been demonstrated their ability to infer potential DDIs rapidly and cheaply[7].

Most computational methods for DDI prediction are based on machine learning, which can be roughly categorized into four classes: feature-based, similarity-based, network-based, and matrix decomposition-based methods. Feature-based methods generally represent drugs into feature vectors derived from various drug properties, such as chemical structure[8], side-effect[9], etc. For example, Zhang *et al.* [10] collected a variety of drug features, i.e., pathway, indication, transporter, etc., and build prediction models respectively by the neighbor recommender method, the random walk method, and the matrix perturbation method. Chu *et al.* [11]constructed an autoencoder to predict potential DDIs. Similarity-based methods assume that two similar drugs tend to have common interactions. They usually calculate the similarity between drugs and represent drugs by similarity vectors [12][13], then predict DDIs via different prediction models. For instance, Gottlieb *et al.* [14] used 7 types of drug features to generate similarity vectors and constructed a DDI prediction model by logistic-based regression. Ferdousi *et al.* [15] utilized structure similarity vectors as the input of a deep neural network to predict potential DDIs. By constructing the knowledge graph of DDIs, network-based methods generally obtain drug embeddings by their local or global topological information in the graph to infer potential DDIs. For example, Wang *et al.* [16] generated drug embedding representation on the drug structure graph by graph convolution neural network (GCN). Huang *et al.* [17] utilized GCN to generate the first and second-order neighbors to represent drugs and predicted DDIs by a coefficient matrix. Wang *et al.* [18] embedded the DDI knowledge graph with comprehensive biomedical text into a common low dimensional space and computed the DDI information through a link prediction process. Takeda *et al.* [19] represented drugs by the position similarity in the DDI knowledge graph and predicted DDIs by the logistic regression model. Matrix factorization-based methods decompose the adjacency matrix of DDIs into several factor matrices and reconstruct the adjacency matrix to identify potential DDIs [20][21][22]. Yu *et al.* [23] predicted potential DDIs through non-negative matrix factorization. Rohani *et al.* [24] integrated a nonlinear multi-similarity fusion with matrix factorization to identify potential DDIs.

Although existing methods have achieved inspiring prediction, some issues are still challenging as follows.

- Inductive prediction. Practical applications usually require two inductive scenarios of DDIs prediction [25] to deduce interactions between new drugs and known drugs or interactions among new drugs. However, many methods, such as network-based and matrix decomposition-based methods, cannot meet the demands in the inductive scenario [26].

- Multi-label interaction prediction. Our observation shows that two drugs may trigger one or more types of drug interactions, however, most existing methods only predict whether two drugs trigger an interaction or trigger a specific type of interaction.

- Interpretability. A predictive model with good interpretability surely helps clinicians to uncover the underlying mechanism of DDIs. Nevertheless, the interpretability of many existing methods is weak yet. Especially, the methods are derived from knowledge-graph or deep learning.

Recent work has demonstrated that the action of a drug is mainly caused by the actions of its local chemical structures (i.e., substructures) rather than its whole structure [27]. Based on this finding, we assume that the interaction between two drugs is determined by the occurrence of their substructures and their DDI types are determined by the linkages between different substructure sets. As illustrated in Fig. 1, two drugs, denoted as $d_a$ and $d_b$, are represented as a set of substructures with specific chemical meanings respectively. They trigger four types of interactions (i.e., multiple-label interactions), colored by yellow, blue, orange, and green from top to bottom (Fig. 1a). Each type of their interactions is caused by a specific association between two substructure sets derived from the two drugs respectively. For example, the yellow interaction is caused by two pairs of substructures (the pair of $s_1^a$ in $d_a$ and $s_3^b$ in $d_b$ as well as the pair of $s_n^a$ and $s_1^b$). The blue interaction is caused by only one pair of substructures, $s_3^a$ in $d_a$ and $s_2^b$ in $d_b$ (Fig. 1b). Overall, the multiple-to-multiple associations between their substructures determine their interactions. Also, a substructure may attend multiple-type interactions. In the context of DDI, we term these associations between substructures as substructure-substructure interactions (SSIs).

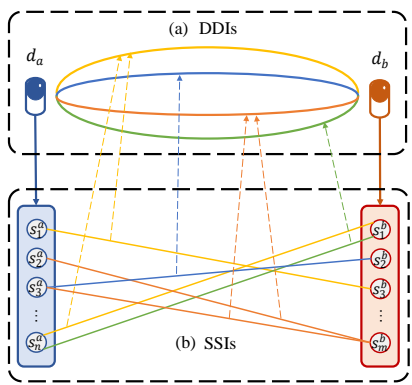

**Fig. 1. SSIs and DDIs.** $(d_a, d_b)$ is a drug pair, $s_i^a$ is the substructure of drug $d_a$, $s_i^b$ is the substructure of drug $d_b$, different colored edges represent the type of different interaction.

Under this assumption, we propose a Substructure-aware Tensor Neural Network for DDI Prediction, referred to as STNN-DDI. The proposed model learns a *substructure×substructure×interaction* tensor (named ***ST***) which characterizes a substructure-substructure interaction (SSI) space, expanded by a series of rank-one tensors [28]. According to a list of predefined substructures with specific chemical meanings (e.g., PubChem fingerprint), two given drugs are embedded into this SSI space to discriminate what types of interactions they trigger and how likely they trigger a specific type of interaction. By leveraging the substructure tensor ***ST***, STNN-DDI has the following advantages.

- It provides a unified form for the DDI prediction in both transductive and inductive scenarios, where the latter accounts for both predicting interactions between new drugs and known drugs and predicting interactions among new drugs.
- It also provides an integrated solution for the case that two drugs trigger single interaction (ordinary multi-type interactions) and the case that they trigger multiple-type interactions simultaneously (multi-label interactions).
- More importantly, its explicit interpretability enables to both reveal an important substructure pair across drugs regarding a DDI type of interest and uncover major interaction type-specific substructure pairs in a given DDI.

The remaining sections are organized as follows. Section 2 briefly introduces the preliminary knowledge of tensors. Section 3 formulates the problem of DDI prediction and gives the detailed modeling and solution of STNN-DDI. Section 4 designs corresponding experiments to demonstrate the superiority of STNN-DDI in different scenarios and illustrate its interpretability. Section 5 concludes our contributions.

## 2 Preliminary

To illustrate STNN-DDI clearly, in this section, we shall briefly review several basic definitions, including tensor and its operations.

A tensor is the generalization of a matrix to >2 dimensions and can consequently be treated as multidimensional fields [28]. Its decompositions originally appeared in 1927 [29], but have remained untouched by the computer science community until the late $20^{th}$ century [30]. Several core terms about tensor are used in this paper.

*Tensor*: An $N$-way or $N^{th}$-order tensor is an element of the tensor product of $N$ vector spaces, each of which has its coordinate system.

*Rank-one tensors*: A $N$-order tensor $\mathcal{Y} \in \mathbb{R}^{I_1 \times I_2 \times \cdots \times I_N}$ is defined as rank-one tensor when it can be strictly decomposed into the outer product of $N$ vectors. $\mathcal{Y}$ is given as:

$$\mathcal{Y} = a^{(1)} \circ a^{(2)} \circ \cdots \circ a^{(N)}, \tag{1}$$

where ($\circ$) is the vector outer product and $a^{(i)}$ is a vector. Each element of tensor $\mathcal{Y}$ is the product of the corresponding vector elements, the operation can be written as:

$$y_{i_1 i_2 \cdots i_N} = a^{(1)}_{i_1} a^{(2)}_{i_2} \cdots a^{(N)}_{i_N}, \tag{2}$$

where $a^{(j)}_{i_j}$ is the $i_j^{th}$ element of $a^{(j)}$.

*Canonical Polyadic (CP) Decomposition*: A $N$-order tensor $\mathcal{X} \in \mathbb{R}^{I_1 \times I_2 \times \cdots \times I_N}$ can be approximately expressed by the sum of rank-one tensors, the operation is defined as:

$$\mathcal{X} \approx \sum_{r=1}^{R} \lambda_r \mathcal{Y}_r, \tag{3}$$

where $\mathcal{Y}_r \in \mathbb{R}^{I_1 \times I_2 \times \cdots \times I_N}$ is a rank-one tensor, $\lambda_r$ is the weight of $\mathcal{Y}_r$. The CP Decomposition of a 3-order tensor $\mathcal{X} \in \mathbb{R}^{I \times J \times K}$ can be expressed as:

$$\mathcal{X} \approx \sum_{r=1}^{R} \lambda_r (a_r \circ b_r \circ c_r) = \langle \lambda; \mathbf{A}, \mathbf{B}, \mathbf{C} \rangle, \tag{4}$$

where $R$ is the rank of tensor $\mathcal{X}$, $\lambda_r$ is the weight of the $r^{th}$ rank-one tensor, and $a_r \in \mathbb{R}^I$, $b_r \in \mathbb{R}^J$, $c_r \in \mathbb{R}^K$ are the vectors decomposed by $\mathcal{X}$ in three directions respectively. Vector $\lambda \in \mathbb{R}^R$ is composed by $\lambda_r$. Factor matrices $\mathbf{A} = [a_1, a_2, \cdots, a_R]$, $\mathbf{B} = [b_1, b_2, \cdots, b_R]$, $\mathbf{C} = [c_1, c_2, \cdots, c_R]$ are the combination of vectors $a, b, c$ in columns where $\mathbf{A} \in \mathbb{R}^{I \times R}$, $\mathbf{B} \in \mathbb{R}^{J \times R}$ and $\mathbf{C} \in \mathbb{R}^{K \times R}$ respectively. These matrices are also called the 2-dimensional representation of $\mathcal{X}$.

## 3 STNN-DDI modeling

In this section, we shall describe STNN-DDI technically, including notations, tensor construction, mathematical principles, and the training by a neural network.

### 3.1 Problem definition

This paper aims to address three tasks of DDIs simultaneously as follows (Fig. 2):

- C1: to infer potential pairwise interactions (marked by dashed lines) between known drugs,
- C2: to infer potential interactions between known drugs having interactions and new drugs,
- C3: to infer potential pairwise interactions among new drugs.

Where we refer to known drugs as the drugs which have reported interactions and new drugs as the drugs which have no known interaction yet. During the construction of a machine learning model, both known drugs and their partial interactions attend the training whereas neither new drugs nor their interaction attends the training. New drugs are only used to test the performance of the trained model in the three abovementioned tasks of DDI prediction.

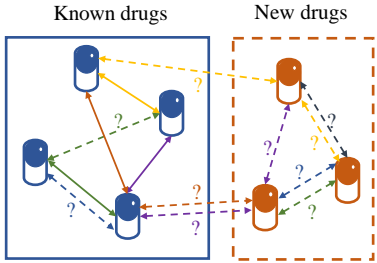

**Fig. 2. Problem definition.** Solid lines represent approved DDIs, and dashed lines represent interactions to be predicted. Different colors of lines represent multiple types of DDIs. The left panel illustrates a graph containing the interactions between known drugs while the right panel shows a graph containing new drugs. Three prediction tasks are marked by C1, C2, and C3 respectively.

**3.2 Notations of STNN-DDI**

Let $S = \{s_i\}_{i=1}^{n}$ be the set of all $n$ substructures or functional groups in a list of predefined substructures with explicit chemical meanings (e.g., PubChem fingerprint), $D = \{d_i\}_{i=1}^{m}$ be the set of $m$ drugs, $L = \{l_i\}_{i=1}^{f}$ be the set of all $f$ relations between drug pairs. Here, $l_k$ is encoded by a one-hot vector $\bm{v}_k = (v_i^k)_{i=1}^{f}$ and $d_p$ is encoded by Pubchem fingerprint, namely $\bm{e}_p = (e_i^p)_{i=1}^{n}$.

Table **1** list the notations of STNN-DDI used in the following sections.

**Table 1.** Notations of STNN-DDI

| | |
|---|---|
| $n$ | $n$=881, which is the number of all the chemical substructures in the PubChem fingerprint |
| $m$ | $m$=555, which is the number of drugs in the dataset |
| $f$ | $f$=1318, which is the number of all types of interactions in the dataset |
| $s_i$ | The $i^{th}$ chemical substructure or functional group in $S$ |
| $d_p$ | The $p^{th}$ drug in $D$ |
| $\bm{e}_p$ | A vector, fingerprint of drug $d_p$ |
| $e_i^p$ | The $i^{th}$ element of $\bm{e}_p$, if $d_p$ contains $s_i$, $e_i^p = 1$, else $e_i^p = 0$ |
| $l_k$ | The $k^{th}$ relation (interaction type) in $L$ |
| $\bm{v}_k$ | A one-hot vector, the embedding of the relation $l_k$ |
| $v_i^k$ | The $i^{th}$ element of $\bm{v}_k$, if $i = k$, $v_i^k = 1$, else $v_i^k = 0$ |
| $DDI_{pqk}$ | Represents the triplet $\langle d_p, d_q, l_k \rangle$, which has the relation $l_k$ in drug pair $(d_p, d_q)$ |
| $P_{pqk}^d$ | Probability of $DDI_{pqk}$ |
| $SSI_{ijk}$ | Represents the triplet $\langle s_i, s_j, l_k \rangle$, which is the relation $l_k$ in $(s_i, s_j)$ |
| $P_{ijk}^s$ | Probability of $SSI_{ijk}$ |
| $\bm{ST}$ | substructure-substructure-interaction tensor |
| $ST_{ijk}$ | The value of the element of the $i^{th}$ row, $j^{th}$ column, and $k^{th}$ dimension in $\bm{ST}$ |
| $\widehat{P_{pqk}^d}$ | estimated value of $P_{pqk}^d$ calculated by STNN-DDI |

### 3.3 Tensor construction in STNN-DDI model

The underlying assumption of STNN-DDI is that the interaction between two drugs is determined by the occurrence of their substructures and their DDI types are determined by the linkages between different substructure sets. Thus, we model DDI from the perspective of substructure-substructure interaction (SSI) by $P_{pqk}^d$, which is defined as:

$$P_{pqk}^d = \sum_{i=1}^{n}\sum_{j=1}^{n} P_{ijk}^s e_i^p e_j^q \tag{5}$$

It represents the occurring probability of the triplet $\langle d_p, d_q, l_k \rangle$ as the sum of the occurring probability of $\{\langle s_i, s_j, l_k \rangle\}$, in which $s_i$ and $s_j$ are the substructures included in drug pair $(d_p, d_q)$.

Following the above definition, by calculating the probability of each SSI under a list of predefined chemical substructures, potential DDIs can be obtained. Thus, an $substructure \times substructure \times interaction$ tensor, named *ST*, is designed. The construction and application of $ST \in \mathbb{R}^{n \times n \times f}$ is given in Fig. 3, where coordinate axis *x* or *y* enumerate the set of pre-defined substructures, coordinate axis *z* enumerates the set of all interaction types between substructure pairs, which is referred to as the SSI space. Because of $P_{ijk}^s = ST_{ijk}$, as Fig. 3, the calculation of $P_{pqk}^d$ by *ST* can be rewritten as:

$$P_{pqk}^d = \sum_{i=1}^{n}\sum_{j=1}^{n} ST_{ijk} e_i^p e_j^q \tag{6}$$

This form allows STNN-DDI to predict DDIs in both transductive and inductive scenarios in a unified form because both known drugs and new drugs are mapped into a common SSI space no matter whether a drug has an interaction or not.

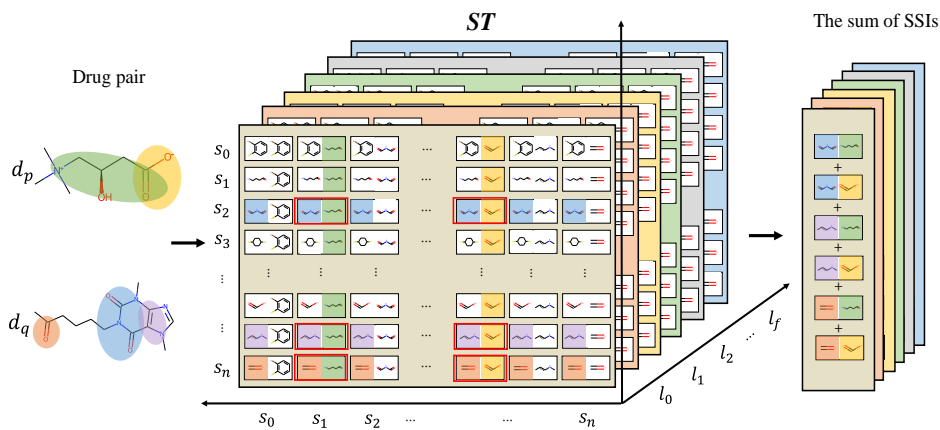

**Fig. 3. Tensor *ST*.** Axis *x* and *y* are substructures and *z* represent the interactions, substructures in the drug pair $(d_p, d_q)$ are highlighted by semitransparent colorful ellipses.

### 3.4 The DDI prediction by tensor *ST*

To perform the DDI prediction via tensor *ST*, we implement Formula (6) by the *n*-mode product for tensor and vectors. Given an *N*-order tensor $\mathcal{X} \in \mathbb{R}^{I_1 \times I_2 \times \cdots \times I_N}$ and a vector $v \in \mathbb{R}^{I_n}$, the *n*-mode product for $\mathcal{X}$ and $v$ is denoted as:

$$\mathcal{X} \bar{\times}_n v_{i_1 \cdots i_{n-1} i_{n+1} \cdots i_N} = \sum_{i_n=1}^{I_n} x_{i_1 \cdots i_N} v_{i_n} \tag{7}$$

Therefore, with the *n*-mode product mathematical principle, $P_{pqk}^d$ can be calculated by *ST* mode product with $v_k$, $e_p$, $e_q$ individually, this operation can be expressed as follows:

$$P_{pqk}^d = ST \bar{\times}_3 v_k \bar{\times}_2 e_p \bar{\times}_1 e_q \tag{8}$$

The mathematical principles are given as follows and shown in Fig. 4.

First, $ST_{::k} = ST \bar{\times}_3 v_k$, where the frontal slices $ST_{::k}$ can be described as an $substructure \times substructure$ matrix for interaction $l_k$.

Then, $st_{(p,k)} = ST_{::k} \bar{\times}_2 e_p$, the $j^{th}$ element of the vector $st_j^{(p,k)} = \sum_{i=1}^{n} ST_{ijk} e_i^p$ can be described as the probability of interaction $l_k$ happens caused by $s_j$ under the action of $d_p$.

Finally, $P_{pqk}^d = st_{(p,k)} \bar{\times}_1 e_q = \sum_{j=1}^{n} st_j^{(p,k)} e_j^q$, which can also be written as $P_{pqk}^d = \sum_{i=1}^{n} \sum_{j=1}^{n} ST_{ijk} e_i^p e_j^q$, $P_{pqk}^d$ can be calculated according to the structure of $d_q$.

### 3.5 The factor matrices of $ST$

According to Eq. (4), a tensor can be approximated by the factor matrices of CP decomposition. Therefore, $ST$ can be represented as factor matrices $A, B \in \mathbb{R}^{n \times R}$ and $C \in \mathbb{R}^{f \times R}$, where $R$ is a manually parameter, represents the number of rank-one tensor decomposed by $ST$.

In CP decomposition, each factor matrix collects the latent information of a specific dimension. $A$ collects the latent information of axis $x$, $B$ collects the latent information of axis $y$, $C$ collects the latent information of axis $z$. Therefore, each row of matrix $A \in \mathbb{R}^{n \times R}$ and $B \in \mathbb{R}^{n \times R}$ can be regard as the embedding of each substructure. Hence, there is $A=B$. While matrix $C \in \mathbb{R}^{f \times R}$ can be regard as the embedding of the relations. Then $ST$ can be expressed as:

$$ST \approx \sum_{r=1}^{R} \lambda_r (a_r \circ a_r \circ c_r) = \langle \lambda; A, A, C \rangle \quad (9)$$

As mentioned in section 3.4, $P_{pqk}^d$ can be calculated by $n$-mode product for $ST$ and $v_k$, $e_p$, $e_q$, and $ST$ can be approximated by the factor matrices $A$ and $C$.

Then, to predict DDIs by these factor matrices, Multilinear Tensor Transformation is introduced [31]. A tensor $\mathcal{X}$ can be transformed on multiple dimensions by hitting each vector producing $\mathcal{X}$ with a transformation vector (or matrix) from the left side. In the 3-dimensional case, the multilinear tensor transformation using vectors $v_i$ is defined as:

$$\mathcal{X} \bar{\times}_1 v_1 \bar{\times}_2 v_2 \bar{\times}_3 v_3 \approx \sum_{r=1}^{R} \lambda_r (a_r \circ b_r \circ c_r) \bar{\times}_3 v_3 \bar{\times}_2 v_2 \bar{\times}_1 v_1 = \sum_{r=1}^{R} \lambda_r (v_1^T \cdot a_r)(v_2^T \cdot b_r)(v_3^T \cdot c_r), \quad (10)$$

where $(\cdot)$ is the dot product between vectors. Therefore, $P_{pqk}^d$ can be calculated as:

$$P_{pqk}^d \approx \sum_{r=1}^{R} \lambda_r (a_r \circ b_r \circ c_r) \bar{\times}_3 v_k \bar{\times}_2 e_p \bar{\times}_1 e_q = \sum_{r=1}^{R} \lambda_r (e_q^T \cdot a_r)(e_p^T \cdot b_r)(v_k^T \cdot c_r)$$
$$= [(e_q^T \cdot A) \odot (e_p^T \cdot A) \odot (v_k^T \cdot C)] \cdot \lambda, \quad (11)$$

where $\odot$ is the Hadama product.

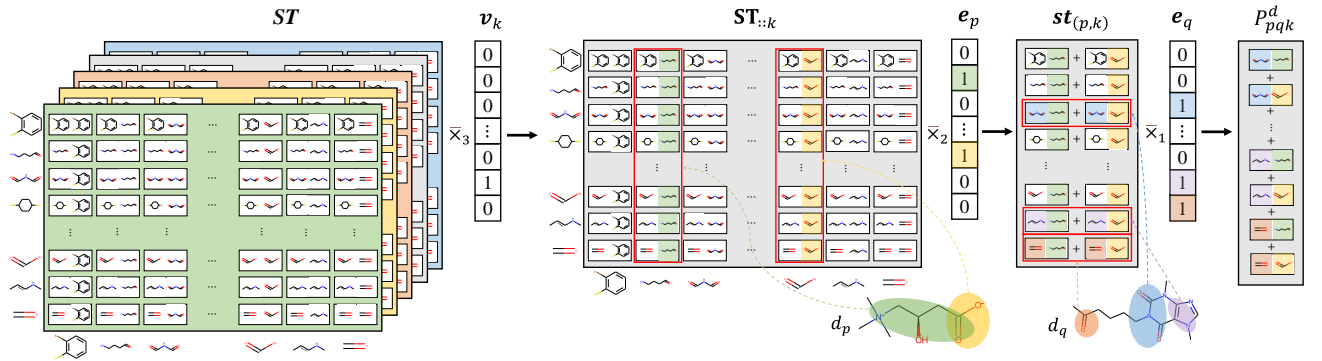

**Fig. 4.** $ST$ mode product with $v_k$, $e_p$, $e_q$ to calculate $P_{pqk}^d$, substructures in the drug pair $(d_p, d_q)$ are highlighted by semitransparent colorful ellipses.

## 3.6 Estimate $P_{pqk}^d$ by neural network

With the above mathematical derivation, $P_{pqk}^d$ can be approximately calculated by the factor matrices of $ST$. Thus, it is reasonable for STNN-DDI to obtain $P_{pqk}^d$ by a neural network. The modeling of STNN-DDI can be described as follows.

Firstly, we initialize $R$-dimension embeddings for every chemical substructure and interaction randomly, as mentioned in section 3.4, $ST$ can be represented by factor matrices **A** and **C**, where $\mathbf{A} \in \mathbb{R}^{n \times R}$ is constituted by the embeddings of all chemical substructures, $\mathbf{C} \in \mathbb{R}^{n \times R}$ is constituted by the embeddings of all interactions. According to Eq. (11), we can get:

$$\widehat{P_{pqk}^d} = [(e_q^T \cdot \mathbf{A}) \odot (e_p^T \cdot \mathbf{A}) \odot (v_k^T \cdot \mathbf{C})] \cdot \mathbf{W}_\lambda + \boldsymbol{bias}, \qquad (12)$$

where $\mathbf{W}_\lambda \in \mathbb{R}^R$ is the weight of each rank-one tensor in tensor reconstruction, $\boldsymbol{bias} \in \mathbb{R}^R$ is added for enhancing the robustness of STNN-DDI. Naturally, the construction of a neural network model is based on the derivation in the above sections. Fig. 5 shows the workflow of the learning process by using a neural network.

Since STNN-DDI has trainable parameters, including **A**, **C** and $\mathbf{W}_\lambda$, we calculate the loss in all classes over all the training subjects by:

$$loss = \sum_{DDI_{pqk} \in DDI_{train}} \left( P_{pqk}^d - \widehat{P_{pqk}^d} \right)^2, \qquad (13)$$

where $P_{pqk}^d$ indicates DDIs if $DDI_{pqk}$ occurs, $P_{pqk}^d = 1$, otherwise 0.

## 4 Experiment

### 4.1 Dataset and experimental setup

The dataset of DDI we used was taken from Zhu *et al.*[32], which collected 555 drugs and their 576,513 pairwise interactions involving 1,318 interaction types from TWOSIDES Tatonetti *et al.*[7]. The original data of TWOSIDES was collected from the source of FARES [33] and MedEffect [34] respectively. More details can be found in TWOSIDES.

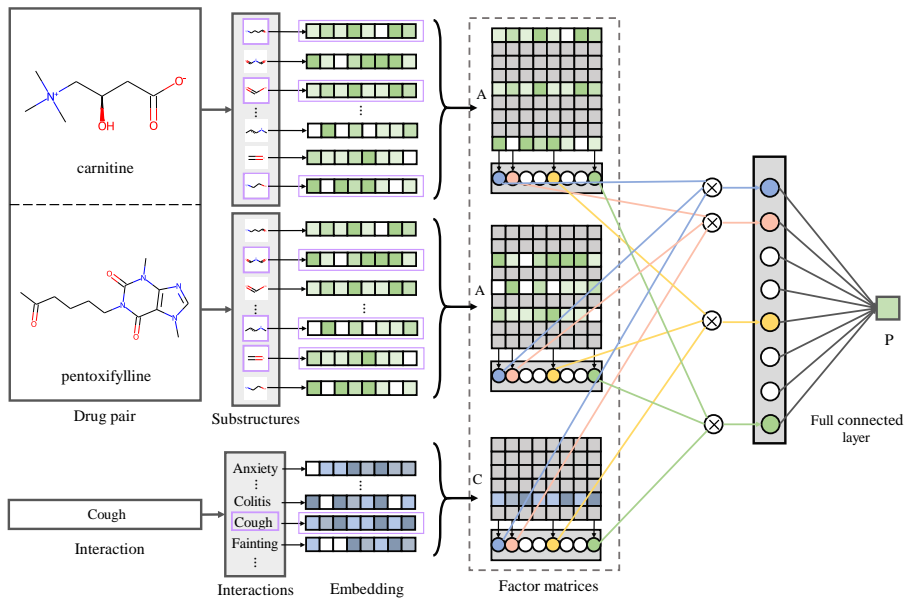

**Fig. 5. The workflow of STNN-DDI.** 1. Input triplet; 2. Initialize $R$-dimensional embedding for all substructures and interactions randomly; 3. The embeddings of substructure and interaction constitute **A** and **C** in rows respectively; 4. Construct a neural network according to Eq. (12) and (14) to calculate $\widehat{P_{pqk}^d}$

We adopted 10-fold cross-validation to estimate the performance of STNN-DDI and compare it with four state-of-the-art methods under three scenarios.

For task C1, all the DDIs were randomly divided into 10 parts of approximately equal sizes in each round, where 90% DDIs are sampled as the training set and the remaining 10% DDIs are used for testing.

For task C2, all the drugs were randomly divided into 10 parts, where 90% of drugs were sampled as known drugs for the training, and the remaining 10% of drugs were treated as new drugs for the testing. Note that none of the interactions of new drugs attend in the training. C2 aims to deduce potential interactions between the training drugs and the testing drugs.

The division of drugs in Task C3 is similar to that in Task C2. Remarkably, task C3 aims to infer potential interactions among the testing drugs by the model learned from the interactions of the training drugs.

In addition, since the number of negative samples is larger than that of positive samples, the class imbalance would impact the predicting performance of the model. Therefore, we use all the positive samples and randomly select the same number of negative samples during the training.

Four evaluating metrics were adopted to measure the models' performance, including AUC, AUPR, Accuracy (Acc), and Precision (Pre).

### 4.2 Hyper-Parameter tuning

The only hyper-parameter of STNN-DDI is the rank of **ST**, denoted as *R*, which is the number of rank-one tensors decomposed by **ST** and indicates the embedding dimension of the SSI space. To investigate how it influences the performance of STNN-DDI, we tuned the value of *R* for the list of {10, 20, 50, 100, 200, 300, 350, 400, 450, 500} under three prediction tasks. The results show that the increment of R brings the better performance of STNN-DDI and achieves the best performance at 400 regarding all four metrics (Fig. 6). In addition, it is observed that the small value of *R* still has an acceptable performance. Thus, STNN-DDI is a robust model to the hyper-parameter. In all the subsequent experiments, we fixed R=400 to run STNN-DDI.

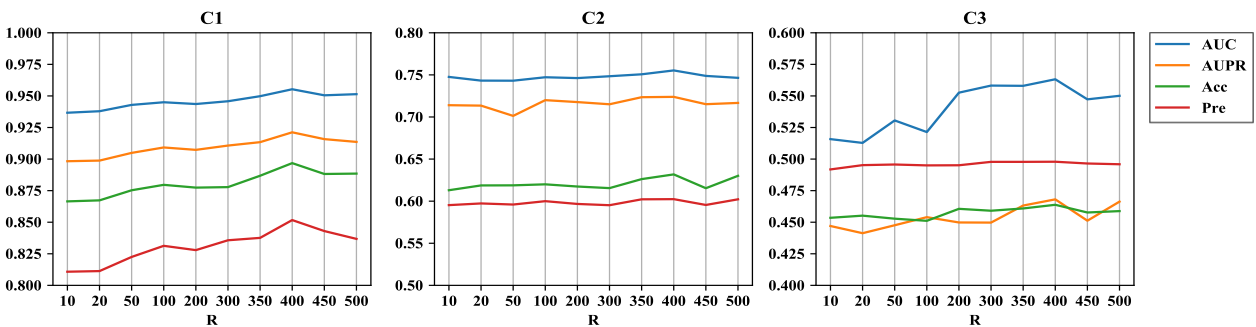

**Fig. 6.** Experimental results of different prediction tasks under different *R*.

### 4.3 Baselines and comparison

To test the performance of STNN-DDI, we compared it with four state-of-the-art baselines, which are briefly introduced as follows.

• DDIMDL [13]: DDIMDL generates drug embeddings through multiple drug structural similarities and trains the model through auto-encoder and neural network.

• DNN [35]: It used a deep learning model to predict DDIs and Deng *et al*. [13] modified it which takes drug molecular fingerprints as input.

• GoGCN [16]: GoGCN extracts drug features in drug structure graphs as well as a DDI knowledge graph by a multi-resolution architecture, and trains the model by a dual attention mechanism.

• Rescal [36]: We also implement a model based on Rescal [36], which is an efficient tensor decomposition model for link prediction. In this model, we organized all the DDIs into a tensor, which was decomposed by Rescal into a core tensor and drug latent representations of drugs, and predict potential interactions by reconstructing the tensor.

Table 2. The experimental results of different models

| Task | Model | AUC | AUPR | Acc | Pre |
|---|---|---|---|---|---|
| C1 | DDIMDL | 0.8060 | 0.7519 | 0.3998 | 0.5382 |
|  | Rescal | 0.9017 | 0.9113 | 0.8828 | 0.8290 |
|  | DNN | 0.9235 | 0.7595 | 0.4163 | 0.538 |
|  | GoGCN | 0.7774 | 0.7103 | 0.7076 | 0.6331 |
|  | STNN-DDI | **0.9553** | **0.9212** | **0.8968** | **0.8517** |
| C2 | DDIMDL | 0.7622 | 0.1568 | 0.7672 | 0.0212 |
|  | DNN | **0.7885** | 0.0243 | **0.7600** | 0.0499 |
|  | STNN-DDI | 0.7553 | **0.7239** | 0.6318 | **0.6024** |
| C3 | DDIMDL | **0.7478** | 0.2404 | 0.7072 | 0.0203 |
|  | DNN | 0.7373 | 0.0280 | **0.7574** | 0.020 |
|  | STNN-DDI | 0.5632 | **0.4681** | 0.4638 | **0.4979** |

The best results of all methods are highlighted in bold.

In addition, since neither GoGCN nor Rescal can be applied in the inductive scenario, we didn't run them in C2 and C3.

The comparison results are shown in Table 2. In a nutshell, STNN-DDI significantly outperforms all the baselines across all the four metrics in task C1. In the case of C2 and C3, STNN-DDI achieves the best in terms of AUPR and Precision, though it cannot beat the baselines in terms of AUC and Acc. Our extra investigation of the underlying reason reveals that 5~10 substructures occurring in the testing drugs don't occur in the training drugs. The SSI space trained without these substructures failed to characterize new drugs and resulted in the worse performance of STNN-DDI in C2 or C3, while STNN-DDI beat all the baselines in C1 due to the SSI space was trained by all the occurring substructures in the drugs. We believe that STNN-DDI would achieve a better performance in C2 and C3 if all the substructure pairs can attend its training (e.g., using pre-train strategies). In summary, the comparison with baselines validates the superiority of STNN-DDI for multi-type DDI prediction in both transductive and inductive scenarios.

### 4.4 Interpretability illustration

We elaborated on two experiments to illustrate the explicit interpretability of STNN-DDI. The first experiment illustrated how STNN-DDI can reveal an important substructure pair across drugs regarding a DDI type of interest, while the second one illustrated how it can uncover major interaction type-specific substructure pairs in a given DDI.

#### 4.4.1 Case study 1: important substructure pairs regarding single interaction type

**ST** can be regarded as $f$ slices of $substructure \times substructure$ matrices, of which each corresponds to a specific type of interaction. In such a matrix, the value of an element indicates how much a pair of corresponding substructures contributes to the interaction type. As we found, it is usual that a few elements are significantly larger than others in the matrix. Thus, substructure pairs having large values are important to the interaction type. In other words, we consider that they determine or cause the interaction type of interest.

The "temperature increased" interaction was selected as a case study. Its most important substructure pair (with the probability $P^s_{228,478,1} = 0.69$) is $\langle s_{228}, s_{478} \rangle$, where $s_{228}$ is '>= 1 saturated or aromatic carbon-only ring size 8' and $s_{478}$ is 'S-C=N-[#1]' in terms of PubChem fingerprints. After searching the database, we found two drugs, Docetaxel and Paclitaxel containing substructure $s_{228}$, as well as four drugs containing substructure $s_{478}$, which are Lansoprazole, Omeprazole, Pantoprazole and Rabeprazole. By counting all the eight drug pairs between $s_{228}$-contained drugs and $s_{478}$-contained drugs and their interaction statements in the dataset, we observed that 7 drug pairs leads to the interaction that patients' body temperature increases, except the drug pair <Paclitaxel, Pantoprazole> (Fig. 7). This finding may conclude that the risk of "temperature increased" could be triggered if two drugs containing ($s_{228}, s_{478}$) are taken together. Similar conclusions were also reported in other publications (Rabanser *et al*., 2017). Thus, the detection of important substructure pairs is helpful to uncover why a DDI type of interest occurs.

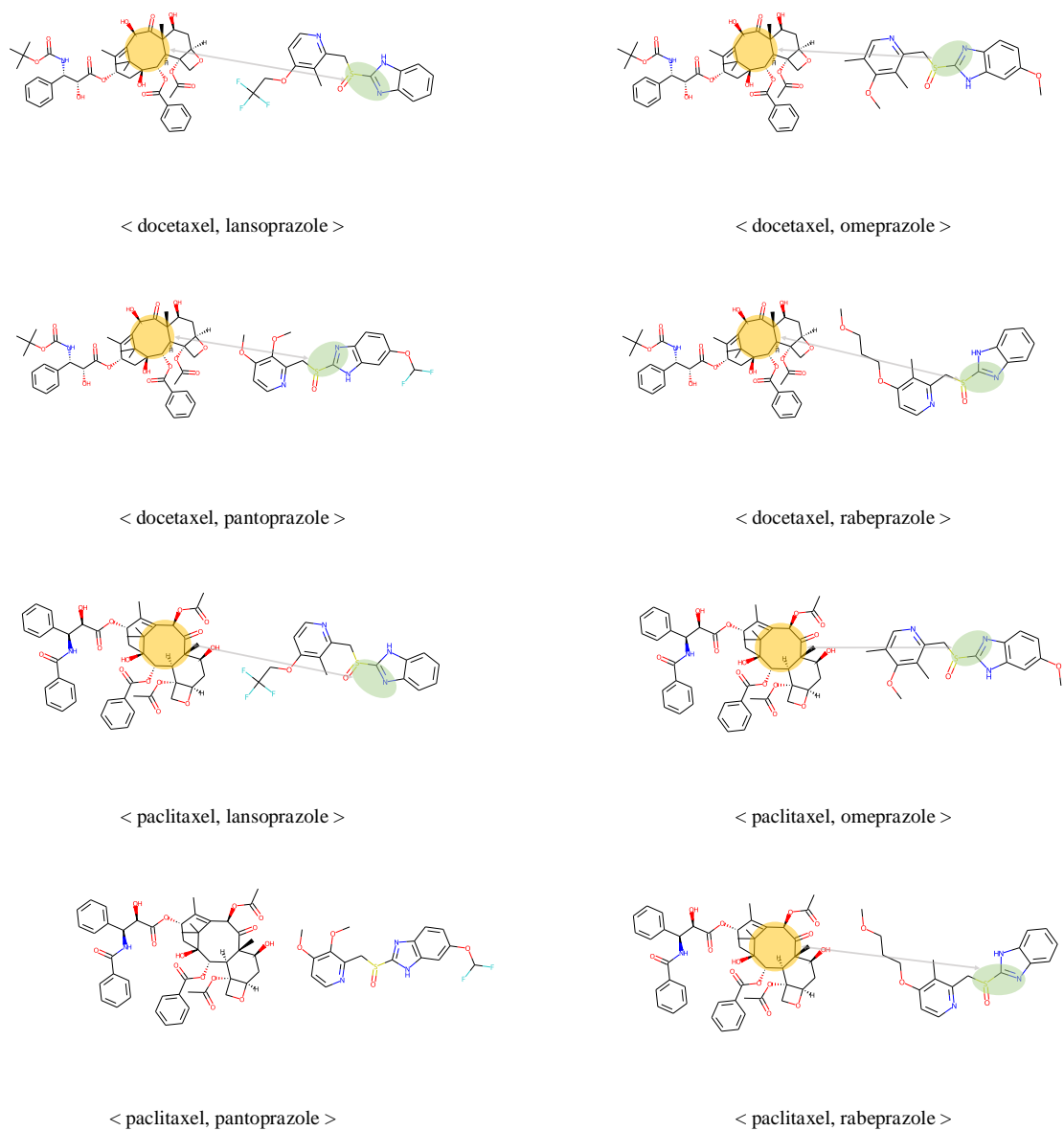

**Fig. 7. Visualization of the important substructure pairs regarding the "temperature increased" interaction.** Only the 8 pairs of drugs(bottom in each panel) contain the substructure pair ($s_{228}$, $s_{478}$), and all the drug pairs can cause patient temperature increased except <Paclitaxel, Pantoprazole> (highlighted by semitransparent colorful ellipses).

### 4.4.2 Case study 2: important substructure pair subsets regarding a drug pair

As we found, many drug pairs trigger more than one type of DDIs, referred to as multi-label DDIs. STNN-DDI can uncover how substructure pairs of two drugs contribute to multi-label DDIs. To illustrate this advantage, we carried out a visualization analysis by following Zhu *et al*. [32] on the occurrence of interactions in a drug pair Carnitine and Budesonide. According to the interaction statements in the dataset, these two drugs trigger 6 types of interactions, including 'blood calcium increased', 'narcolepsy', 'pyoderma', 'apoplexy', 'emesis' and 'abdominal hernia'.

After enumerating all their substructure pairs in each type of their interactions, we picked up important substructure pairs, where a set of substructures coming from Carnitine and another set of substructures coming from Budesonide (Fig. 8). All of them are called 'multiple to multiple' substructure-substructure interactions.

In detail, for 'blood calcium increased', there are 3 pairs between three important substructures <'C(~C)(~C)(~H)(~O)', 'C(-C)(=O)', 'C-C-C-O-[#1]'> from Carnitine and two <'OC1C(O)CCC1', '(C(=O)CO[C@@]'> from Budesonide, marked by ①, ② and ③ respectively (Fig. 8-a). Similarly, there are 4 important substructure pairs associated with 'narcolepsy', 3 associated with 'pyoderma', 3 associated with 'apoplexy', 3 associated with 'emesis' and 4 associated with 'abdominal hernia' (Fig. 8-b, c, d, e, f). Especially, 4 interactions of 'abdominal hernia' involve 4 substructures coming from Carnitine and another 4 substructures coming from Budesonide. Moreover, the two substructure <'OC1C(O)CCC1', '(C(=O)CO[C@@]'> in Budesonide attend all types of its interactions.

This illustration demonstrates that STNN-DDI can uncover interaction type-specific substructure pairs in a given DDI. The exclusive advantage of STNN-DDI, compared with other models, enables the explanation of why two drugs trigger multi-type even multi-label interactions. Its interpretability can provide valuable guidance in the development of new drugs and poly-drug therapy.

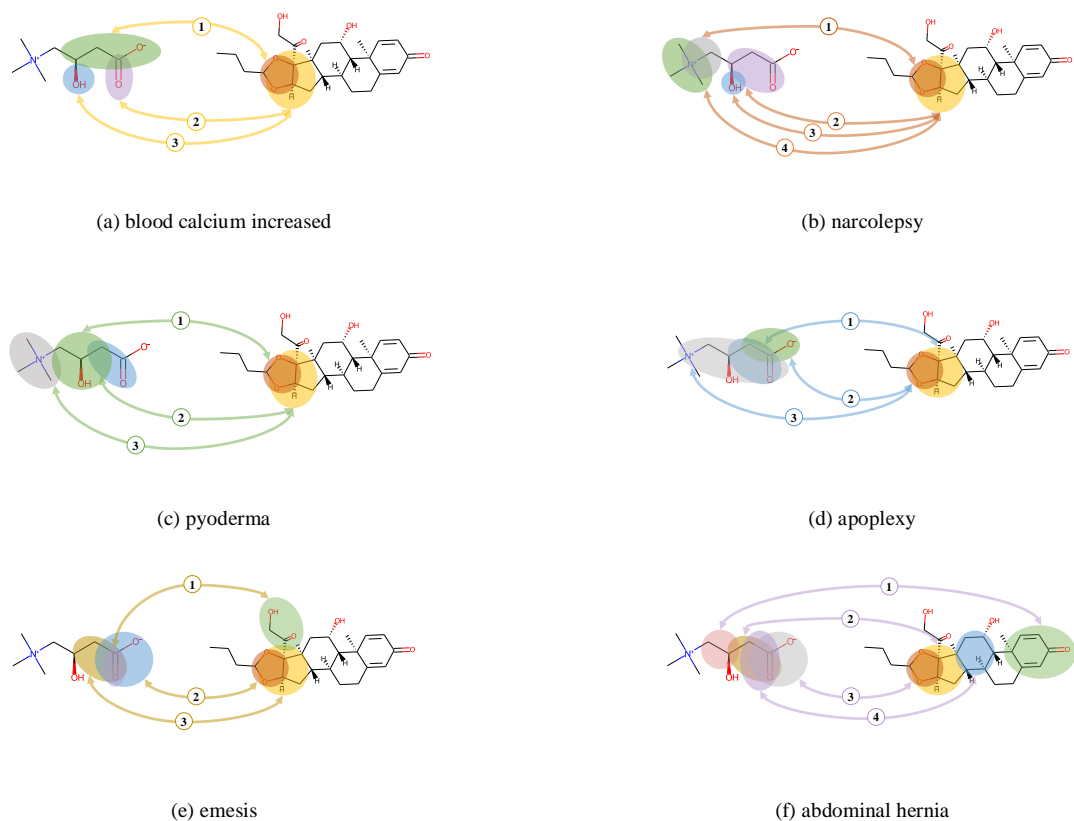

(a) blood calcium increased
(b) narcolepsy
(c) pyoderma
(d) apoplexy
(e) emesis
(f) abdominal hernia

**Fig. 8. Multi-label DDIs between Carnitine and Budesonide.** Six types of interactions(bottom in each panel) between Carnitine (left side in each panel) and Budesonide (right side in each panel) are highlighted by different colors. Important substructures are highlighted by semitransparent colorful ellipses. Curves with double arrows denote important structure pairs, where ①, ②, ③ ④ are their indices of important structure pairs in each of the interaction types.

## 5 Conclusions

In this paper, we've designed a novel end-to-end model, STNN-DDI, for multi-type DDI prediction, based on the assumption that the interaction between two drugs is determined by the occurrence of their substructures and their DDI types are determined by the linkages between different substructure sets. STNN-DDI leverages a 3-D substructure-aware tensor and a fully connected neural network to construct a substructure-substructure interaction (SSI) space, where drugs are embedded to discriminate what types of interactions they trigger and how likely they trigger a specific type of interaction. Due to the SSI space, STNN-DDI is not only a model of transductive DDI prediction but also a model of inductive DDI prediction. Moreover, it can handle the case that two drugs trigger one type of interaction as well as multiple types of interactions simultaneously because the DDI types are determined by the association/interaction between different substructure sets. More importantly, since predefined substructures own specific chemical meanings, it can explicitly reveal an important substructure pair across drugs regarding a DDI type of interest and uncover major interaction type-specific substructure pairs in a given DDI. To summarize, STNN-DDI contributes not only an effective predicting model of multiple-type DDIs but also an explorer to the DDI mechanism.

## Acknowledgments

None

## Availability dataset and source code

The dataset and running code are available at: https://github.com/zsy-9/STNN-DDI.

## Funding

This work has been supported by the National Natural Science Foundation of China (Grant No. 61872297) and Shaanxi Provincial Key Research & Development Program, China (Grand No. 2020KW-063).

*Conflict of Interest:* There is no conflict of interest.